%% file: main.tex
\documentclass[11pt]{article}

\usepackage[letterpaper,margin=1in]{geometry}
\usepackage{setspace}

\usepackage[T1]{fontenc}
\usepackage[utf8]{inputenc}
\usepackage{tgtermes}      
\usepackage{inconsolata}   
\usepackage{microtype}

\usepackage{xcolor}
\definecolor{trnavy}{RGB}{16,66,129}        
\definecolor{trblue}{RGB}{42,120,214}       
\definecolor{linkpurple}{RGB}{140,30,120}   
\definecolor{ourscolor}{RGB}{220,240,220}   
\definecolor{headerblue}{RGB}{16,66,129}    
\definecolor{trgreen}{RGB}{0,100,0}
\definecolor{algcomment}{RGB}{39,100,120}
\definecolor{babyblueeyes}{rgb}{0.63,0.79,0.95}
\definecolor{brightpink}{HTML}{D8315B}
\definecolor{lightpink}{HTML}{EF798A}
\definecolor{cvprblue}{rgb}{0.21,0.49,0.74}
\definecolor{myblue}{HTML}{d8ebf8}
\definecolor{lightred}{HTML}{D33E43}
\definecolor{mygreen}{HTML}{2DD881}
\definecolor{darkgreen}{HTML}{006400}
\definecolor{salmon}{HTML}{FA8072}
\definecolor{mybluee}{RGB}{0,102,204}

\usepackage{amsmath,amssymb,amsfonts,amsthm}
\usepackage{mathtools}
\usepackage{nicefrac}

\usepackage{graphicx}
\usepackage{fontawesome5}   
\usepackage{booktabs}
\usepackage{array}
\usepackage{colortbl}
\usepackage{multirow}
\usepackage{tabularray}
\usepackage{subcaption}
\usepackage{wrapfig}
\usepackage[font=small,labelfont=bf]{caption}

\usepackage{algorithm}
\usepackage[noend]{algpseudocode}

\usepackage{enumitem}
\usepackage{xspace}
\usepackage[normalem]{ulem}

\usepackage[most]{tcolorbox}
\tcbuselibrary{breakable,skins}

\usepackage{titlesec}
\titleformat{\section}{\normalfont\Large\bfseries\color{trnavy}}{\thesection}{0.7em}{}
\titleformat{\subsection}{\normalfont\large\bfseries\color{trnavy}}{\thesubsection}{0.6em}{}
\titleformat{\subsubsection}{\normalfont\normalsize\bfseries\color{trnavy}}{\thesubsubsection}{0.6em}{}

\usepackage{natbib}
\usepackage[colorlinks=true,breaklinks=true,
            linkcolor=linkpurple,citecolor=linkpurple,
            urlcolor=linkpurple,allcolors=linkpurple]{hyperref}
\usepackage{url}

\input{math_commands.tex}

\newtcolorbox{keybox}[1][]{
  enhanced, breakable,
  colback=trblue!5, colframe=trblue!55, boxrule=0.8pt, arc=2.2mm,
  left=8pt, right=8pt, top=6pt, bottom=6pt,
  fonttitle=\bfseries\small\color{white}, coltitle=white,
  colbacktitle=trnavy, attach boxed title to top left={xshift=6pt,yshift=-2.6mm},
  boxed title style={colframe=trnavy,arc=1mm}, #1}

\def\name{\textsc{ThinkRetrieve}\xspace}
\newcommand{\soumya}[1]{} 
\def\*#1{\mathbf{#1}}

\newcommand\blfootnote[1]{\begingroup\renewcommand\thefootnote{}\footnote{#1}\addtocounter{footnote}{-1}\endgroup}

\newcommand{\headerdate}{}   

\title{\name: Retrieval-Augmented Reasoning Traces \\[2pt] for Test-Time Scaling}

\begin{document}

\thispagestyle{empty}
\begin{flushleft}
\begin{minipage}[c]{0.74\textwidth}
  \raisebox{-0.5\height}{\includegraphics[height=11mm]{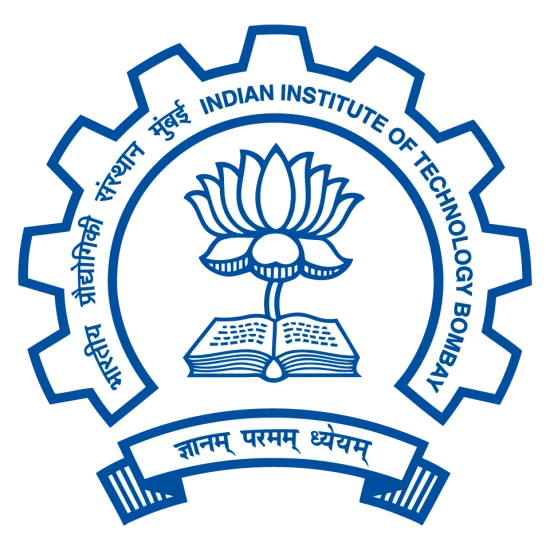}}\hspace{7mm}
  \raisebox{-0.5\height}{\includegraphics[height=11mm]{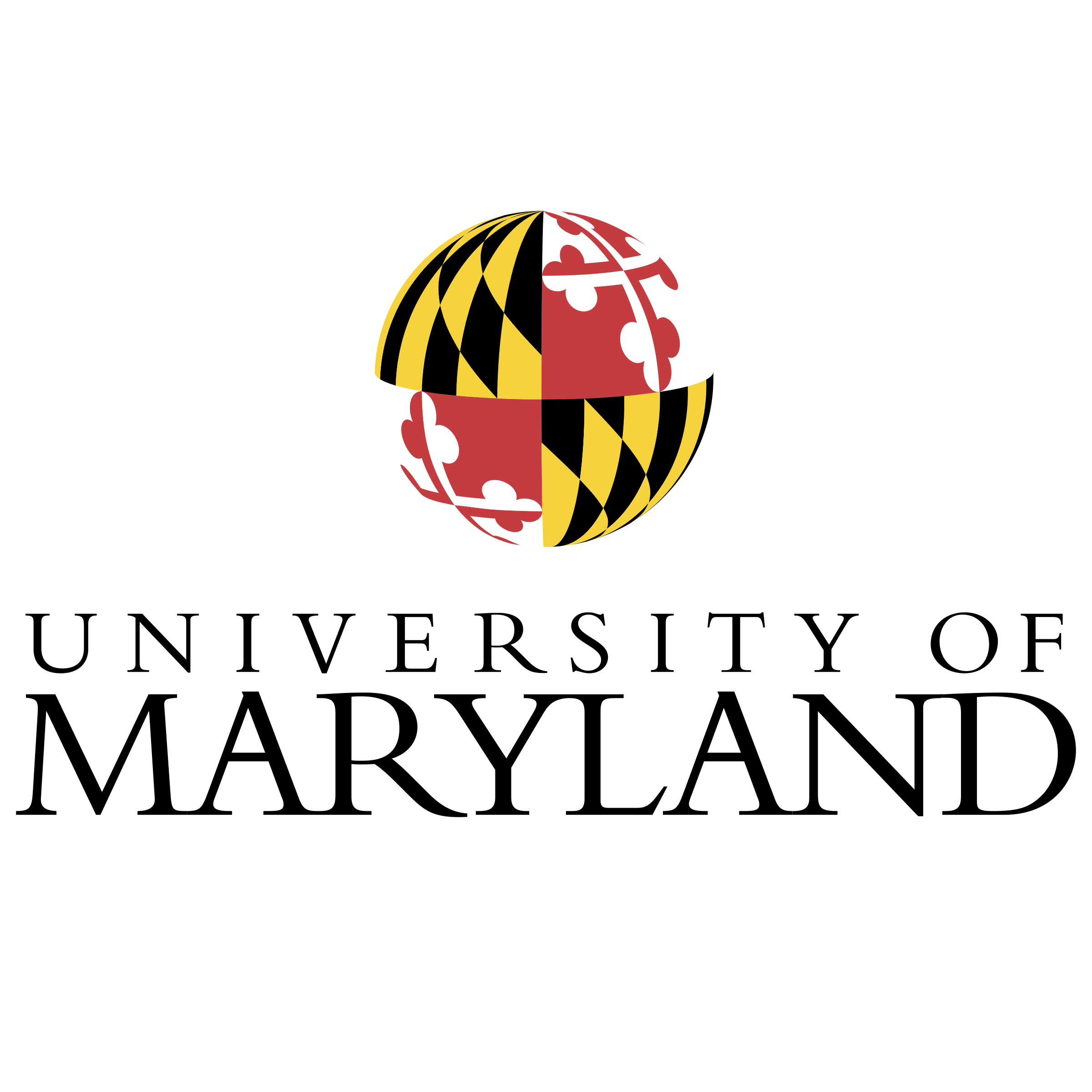}}\hspace{7mm}
  \raisebox{-0.5\height}{\includegraphics[height=6.5mm]{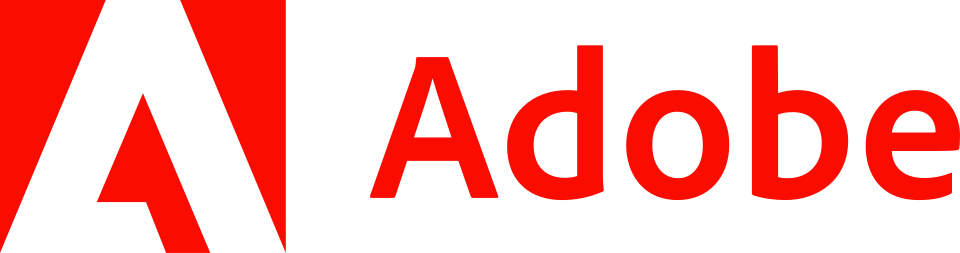}}
\end{minipage}\hfill
\begin{minipage}[c]{0.22\textwidth}\raggedleft\itshape\small \headerdate\end{minipage}
\end{flushleft}
\vspace{2pt}
{\color{trnavy}\rule{\textwidth}{1.2pt}}
\vspace{10pt}

\begin{center}
{\LARGE\bfseries \name: Retrieval-Augmented Reasoning Traces\\[3pt]
 for Test-Time Scaling\par}
\vspace{14pt}

{\large
Vaibhav Singh\textsuperscript{1}\quad
Soumya Suvra Ghosal\textsuperscript{2,3,$\dagger$}\quad
Sarvesh Gharat\textsuperscript{1}\quad
Soumyabrata Pal\textsuperscript{3}\\[3pt]
Ramasuri Narayanam\textsuperscript{3}\quad
Dinesh Manocha\textsuperscript{2}\par}
\vspace{7pt}
{\small\textsuperscript{1}IIT Bombay \quad
 \textsuperscript{2}University of Maryland, College Park \quad
 \textsuperscript{3}Adobe Research\par}
\end{center}
\blfootnote{\hspace*{-1.8em}$^{\dagger}$Most of this work was done while the author was at University of Maryland, College Park.}
\vspace{6pt}

\begin{center}
\begin{minipage}{0.92\textwidth}
\begin{center}{\bfseries\large\color{trnavy}Abstract}\end{center}
\small
Large Reasoning Models (LRMs) improve performance by allocating additional inference-time compute to generate extended chain-of-thought reasoning. However, recent studies reveal that sequential test-time scaling often yields diminishing or even negative returns, as longer traces exhibit increased uncertainty, error compounding, and drift from the original problem. We propose \name, a test-time scaling framework that augments the reasoning traces of LRMs with dynamically retrieved solved examples at each reasoning step. Given an external corpus of problems paired with step-by-step solutions, \name retrieves relevant exemplars at each intermediate step and injects them directly into the thinking trace, providing the model with guidance on \emph{how} to reason rather than merely \emph{what} facts are relevant. Experiments across five reasoning models (1.5B--8B parameters) on GSM-8K, MATH-500, AIME 2025, and SciQ demonstrate that \name consistently improves accuracy over standard test-time scaling, with relative gains of up to $60\%$ on AIME 2025.
\end{minipage}
\end{center}
\vspace{4pt}

\definecolor{iconcolor}{RGB}{216,104,52}  
\begin{center}
\begin{tcolorbox}[enhanced, width=0.92\textwidth,
    colback=trnavy!4, colframe=trnavy!30, boxrule=0.6pt, arc=1mm,
    left=10pt, right=10pt, top=6pt, bottom=6pt, halign=left]
\small
\setlength{\tabcolsep}{6pt}
\begin{tabular}{@{}l l@{}}
{\color{iconcolor}\faGlobe}  & \href{https://itsvaibhav01.github.io/ThinkRetrieve/}{\textcolor{trnavy}{\texttt{https://itsvaibhav01.github.io/ThinkRetrieve/}}} \\[3pt]
{\color{iconcolor}\faGithub} & \href{https://github.com/itsvaibhav01/ThinkRetrieve}{\textcolor{trnavy}{\texttt{https://github.com/itsvaibhav01/ThinkRetrieve}}} \\[3pt]
{\color{iconcolor}\faPython} & \href{https://pypi.org/project/thinkretrieve}{\textcolor{trnavy}{\texttt{pip install thinkretrieve}}} \\
\end{tabular}
\end{tcolorbox}
\end{center}
\vspace{8pt}

\input{section/introduction}

\input{section/related_works}

\input{section/method}

\input{section/experiments}

\input{section/conclusion}

\clearpage
\bibliography{main}

\clearpage
\appendix
\input{section/Appendix}

\end{document}

%% file: math_commands.tex
\usepackage{amsmath,amsfonts,bm}

\def\eqref#1{equation~\ref{#1}}

\def\1{\bm{1}}

\DeclareMathAlphabet{\mathsfit}{\encodingdefault}{\sfdefault}{m}{sl}
\SetMathAlphabet{\mathsfit}{bold}{\encodingdefault}{\sfdefault}{bx}{n}

\DeclareMathOperator*{\argmax}{arg\,max}

%% file: section/introduction.tex
\section{Introduction}
\label{sec:intro}

\begin{figure*}[t]
    \centering
    \includegraphics[width=0.9\textwidth]{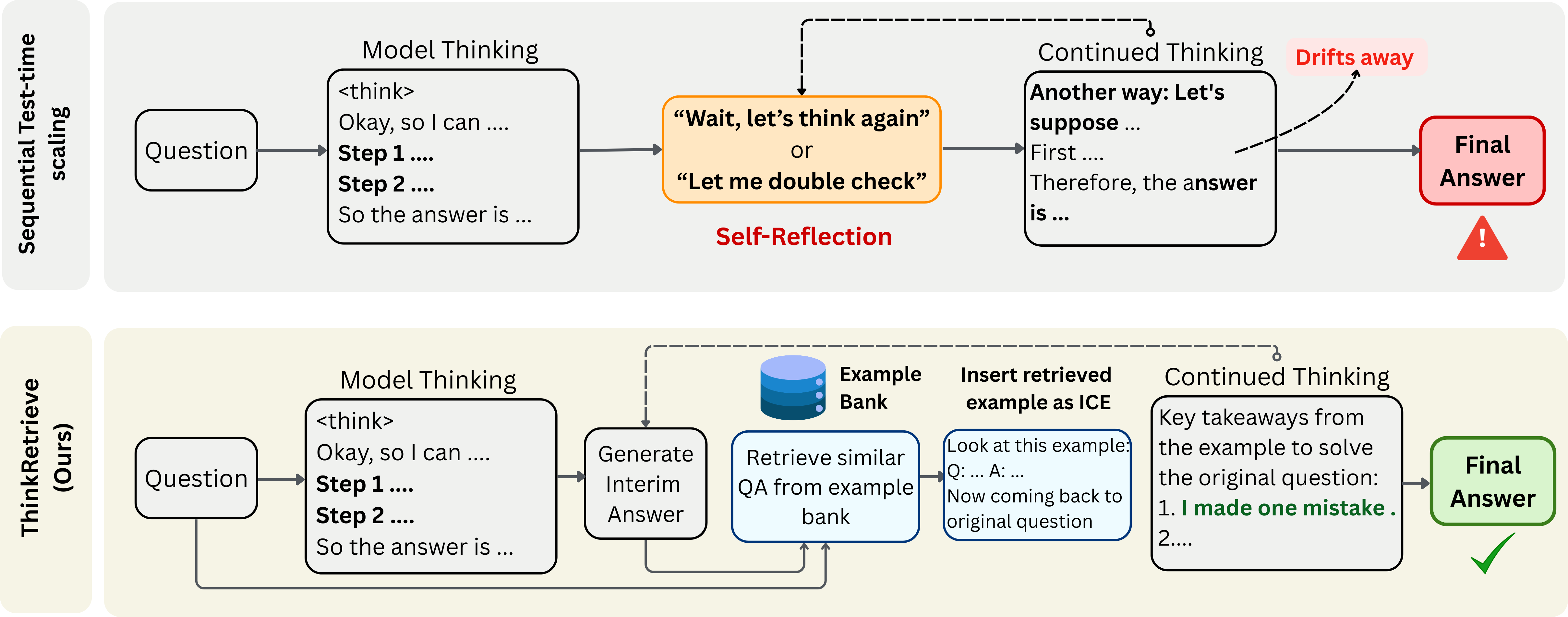}
    \caption{\textbf{Comparison of sequential test-time scaling~\citep{muennighoff2025s1} and \name.} \textbf{(Top)} In standard sequential test-time scaling~\citep{muennighoff2025s1}, the model generates an initial reasoning trace and then relies on self-reflection (e.g., ``Wait, let's think again'') to extend its thinking. However, without access to external guidance, the continued reasoning often drifts away from the correct solution path, leading to an incorrect final answer. \textbf{(Bottom)} \name augments the reasoning process with retrieved exemplars. After the initial thinking phase, the model generates an intermediate answer, which is used as a query to retrieve a relevant solved example from an external example bank. The retrieved exemplar is inserted into the thinking trace as an in-context example (ICE), enabling the model to extract key takeaways, identify errors in its own reasoning, and course-correct before producing the final answer.}
    \label{fig:teaser}
\end{figure*}

Large Reasoning Models (LRMs)~\citep{deepmind2024gemini25, guo2025deepseek, openo1, yang2025qwen3, xai2025grok35} achieve strong performance on multi-step tasks~\citep{cobbe2021gsm8k, hendrycks2021measuring, AIME, zhuo2024bigcodebench, rein2024gpqa} by allocating extra inference-time compute to generate explicit Chain-of-Thought reasoning~\citep{wei2022chain, muennighoff2025s1}. This test-time scaling (TTS) paradigm~\citep{snell2024scaling, muennighoff2025s1}, where performance improves as models ``think longer'', has emerged as a compelling alternative to scaling parameters or pre-training data.

However, recent studies show that thinking longer does not always mean thinking better~\citep{ghosaldoes, gema2025inversescalingtesttimecompute, yue2025don, shojaee2025illusion}: as traces grow, LRMs exhibit increased uncertainty, repetitive cycling, drift, and error compounding~\citep{ghosaldoes, chen2024not, heyman2025reasoning, mirzadeh2024gsm}, and additional compute often amplifies rather than corrects the mistake. This contrasts with how humans tackle hard problems: instead of persisting along a single line of thought, we recall analogous solved problems to guide and verify our reasoning~\citep{gick1980analogical, gentner1983structure, hofstadter2001analogy}.

Existing retrieval-augmented methods do not provide this mechanism for LRMs. Retrieval-Augmented Generation~\citep{lewis2020retrieval, gao2023retrieval} and in-context learning~\citep{brown2020language, luo2023dr, rubin2022learning} both retrieve \emph{once, before reasoning begins}, leaving the retrieved context static throughout generation. Recent dynamic methods such as Search-o1~\citep{li2025search} and RAT~\citep{wang2024rat} interleave retrieval with thinking but target \emph{factual knowledge}, not solved examples that serve as procedural scaffolds.

In this work, we propose \name, a framework that retrieves relevant solved examples and injects them directly into the reasoning trace of LRMs at each thinking step. The core idea of our framework is that, after each reasoning step, the model generates an intermediate answer, which is then used to retrieve a similar solved problem from an external example bank. This retrieved example, containing both the question and its step-by-step solution, is then inserted into the thinking trace before the model continues reasoning. This process repeats at each step, allowing the model to continuously ground its reasoning in concrete worked-out examples, much like how humans recall and reference similar problems when solving a difficult task~\citep{gick1980analogical, hofstadter2001analogy}. Crucially, unlike prior work that retrieves factual documents or knowledge during reasoning~\citep{li2025search, wang2024rat}, \name retrieves {complete solved problems with reasoning chains}, showing the model how to reason rather than what facts to use.

\begin{keybox}[title=Contributions]
\begin{itemize}[leftmargin=1.4em,itemsep=2pt,topsep=2pt]
    \item We propose \name, a framework that
    integrates the retrieval of solved exemplars directly into the reasoning traces of LRMs at each reasoning step, enabling dynamic example-guided test-time scaling.
    \item We provide a thorough analysis showing that in-trace example retrieval significantly reduces the entropy and uncertainty of generated answers compared to
    sequential test-time scaling~\citep{muennighoff2025s1}, offering insight into {why} procedural scaffolding is more effective than simply thinking longer.
    \item We conduct extensive experiments across five reasoning models (DeepSeek-R1-Distill-Qwen-1.5B, Qwen3-1.7B, Qwen3.5-2B, Qwen3-4B, and Qwen3-8B)~\citep{guo2025deepseek, yang2025qwen3} on four benchmarks (GSM-8K~\citep{cobbe2021gsm8k}, MATH-500~\citep{hendrycks2021measuring}, AIME 2025, and SciQ~\citep{welbl2017crowdsourcing}), with two additional retrieval baselines beyond sequential test-time scaling, demonstrating consistent improvements over every baseline across all (model, benchmark) cells.
\end{itemize}
\end{keybox}

%% file: section/related_works.tex
\section{Related Works}
\label{sec:related_works}

\noindent\textbf{Large reasoning models.} OpenAI's o1~\citep{openo1} series introduced the 
first large-scale language model to integrate RL-driven training with the concept of test-time thinking, unlocking the reasoning potential of LLMs and establishing the domain of {large reasoning models} (LRMs)~\citep{xu2025towards}. Following the release of o1, numerous efforts have sought to replicate its reasoning capabilities, including Open-R~\citep{wang2024openr}, Llama-berry~\citep{zhang2024llama}, Journey-R~\citep{qin2024o1}, and Drt-o1~\citep{wang2024drt}. DeepSeek-R1~\citep{guo2025deepseek} was among the first 
open-source models to match o1-level reasoning, employing sparse-reward RL training~\citep{shao2024deepseekmath} on DeepSeek-V3-Base~\citep{deepseekai2024deepseekv3technicalreport}. Subsequently, several open-source efforts have continued employing RL to improve reasoning~\citep{qwq32b, arora2025traininglanguagemodelsreason, aggarwal2025l1, liu2025visual, shen2025vlm, yu2025dapo, huang2025vision, jin2025search, team2025kimi, 
xai2025grok35, deepmind2024gemini25}. In this work, we focus not on improving the training of reasoning models, but on enhancing their test-time behavior through retrieval-augmented reasoning traces.

\vspace{0.1cm}
\noindent\textbf{Test-time scaling in reasoning language models.} 
\citet{snell2024scaling} established that scaling inference-time compute can yield 
predictable performance gains, providing a theoretical grounding for the test-time 
scaling paradigm. \citet{muennighoff2025s1} introduced budget forcing to replicate 
the test-time scaling behavior observed in o1~\citep{openo1}, demonstrating that 
controlling reasoning length at inference can improve performance without 
additional training. L1~\citep{aggarwal2025l1} proposed length-controlled policy 
optimization for precise control over reasoning trace length. 
\citet{yang2025thinkingoptimalscalingtesttimecompute} introduced thinking-optimal 
scaling, training models to adapt dynamically to different reasoning effort levels 
depending on the compute budget. A growing body of work has also explored training 
models to reason efficiently according to task 
complexity~\citep{arora2025traininglanguagemodelsreason, fang2025thinkless, 
zhang2025continue, jiang2025think, liang2025thinkswitcher, zhang2025adaptthink, 
huang2025adactrl}. However, these approaches scale compute by generating {longer} or {more adaptive} reasoning traces, without modifying the {content} available to the model during reasoning. Our work is complementary: rather than controlling how much the model thinks, we enrich what it thinks by injecting retrieved solved examples into the reasoning trace.

\vspace{0.1cm}
\noindent\textbf{Retrieval-augmented reasoning.} Retrieval-Augmented Generation (RAG)~\citep{lewis2020retrieval} augments LLMs with external knowledge retrieved at the input level, and has been widely adopted for knowledge-intensive tasks~\citep{gao2023retrieval}. In the context of reasoning, several recent works have explored integrating retrieval into the reasoning process itself. 
Search-o1~\citep{li2025search} enables LRMs to autonomously trigger search queries during reasoning to address knowledge gaps, while RAT~\citep{wang2024rat} iteratively revises each thought step with retrieved 
information. RARE~\citep{wang2025rare} decouples knowledge storage from reasoning by externalizing domain knowledge to retrievable sources during training. 
Chain-of-Retrieval~\citep{wang2025chainofretrieval} interleaves retrieval with
generation steps for multi-hop question answering. On the in-context learning
front, retrieval of demonstrations has been explored for improving few-shot prompting~\citep{ghosal2025promptrefine, rubin2022learning, luo2023dr}, but such retrieval is performed once before generation and remains static throughout inference. More recent step-aware retrieval methods retrieve premises, per-step process signals, or induced guidelines~\citep{retrievalprm, realprover, guidelineforest}, and math-specific retrieval benchmarks~\citep{mathnetretrieve} show that structurally faithful retrieval over math problems is challenging with off-the-shelf encoders. In contrast, our work retrieves \emph{complete solved problems with step-by-step reasoning trajectories} and injects them at each step of the thinking trace, providing guidance on how to reason rather than supplementing what the model knows.

%% file: section/method.tex
\section{Methodology}
\label{sec:methodology}

\subsection{Preliminaries}
\label{sec:preliminaries}

\noindent\textbf{Mathematical formulation of the reasoning process.} 
We consider Large Reasoning Models (LRMs) that produce explicit intermediate reasoning before generating a final answer. Formally, given a textual input prompt $x = (x_1, x_2, \ldots, x_P)$ consisting of $P$ tokens over a vocabulary $\mathcal{V}$, the reasoning process can be expressed as:
\begin{equation}
    x \rightarrow z \rightarrow y,
    \label{eq:thinking}
\end{equation}
where the model first produces a reasoning trace $z \sim \pi_\theta(\cdot \mid x)$ and subsequently generates the final answer $y \sim \pi_\theta(\cdot \mid x, z)$, with $\pi_\theta$ denoting the language model parameterized by $\theta$.

\noindent\textbf{Test-time scaling via sequential self-reflection.} 
Following~\citet{muennighoff2025s1}, test-time scaling (TTS) extends the thinking 
process by decomposing the reasoning trace into sequential steps:
\begin{equation}
    x \rightarrow z_1 \rightarrow z_2 \rightarrow \cdots 
    \rightarrow z_k \rightarrow y,
    \label{eq:tts}
\end{equation}
where, given the input $x$, the model first generates an initial reasoning step $z_1 \sim \pi_\theta(\cdot \mid x)$. Rather than producing the final answer immediately, the model is prompted to continue reasoning using special instruction tokens (\emph{e.g.}, 
``Wait'', ``Think more''), denoted by $c$. Subsequent reasoning steps are 
sampled iteratively as $z_t \sim \pi_\theta(\cdot \mid x, 
z_{1:t-1}, c)$ for $t = 2, \ldots, k$, until the model generates the final 
response as $y \sim \pi_\theta(\cdot \mid x, z_{1:k}, c)$. For brevity, we omit explicit mention of $c$ in the conditioning and denote 
the thinking traces as $y \sim \pi_\theta(\cdot | x, z_{1:k})$.

\subsection{\name: Retrieval-Augmented Test-Time Scaling}
\label{sec:thinkretrieve}

The core idea behind \name is to transform the sequential test-time scaling process (Eq.~\ref{eq:tts}) into a retrieval-guided reasoning framework. At a high level, after each reasoning step, we prompt the model to generate an intermediate answer reflecting its current solution estimate. This intermediate answer, together with the original query, is used to retrieve a structurally relevant solved example from an external corpus. The retrieved example is then injected into the thinking trace as an exemplar, guiding the model's subsequent reasoning step. This produces an interleaved reasoning trajectory that tightly couples the model's evolving reasoning state with externally grounded solution strategies. We describe each component below.

\noindent\textbf{Example corpus.} For retrieval, our framework leverages an external corpus of solved examples $\mathcal{E} = \{(q_j, a_j)\}_{j=1}^{N}$, where each entry pairs a query $q_j$ with its corresponding step-by-step 
solution $a_j$. For this study, we populate this corpus using an existing collection of synthetically generated questions and solutions, ensuring zero overlap with our evaluation benchmarks (see Section \ref{sec:experiment} for details). Each example is represented by a dense embedding of its query-answer pair $\mathbf{e}_j = \text{Enc}(q_j, a_j)$ using a pre-trained 
sentence encoder $\text{Enc}(\cdot)$, and all embeddings are indexed for efficient nearest-neighbor retrieval.

\subsubsection{Retrieval-augmented thinking} 

Given a test query $x_{\text{test}}$ and a thinking budget of $B$ tokens, \name proceeds iteratively, constructing an interleaved reasoning trajectory $\tau_k = (z_1, e_1, z_2, e_2, \ldots, z_k, e_k)$, where $z_t$ denotes the reasoning step generated by the model at step $t$, $e_t$ denotes the exemplar retrieved after $z_t$, and $k$ is the total number of reasoning steps determined by the token budget $B$.

\noindent\textbf{Step boundary.} A step boundary is triggered when the model naturally produces the \texttt{</think>} stop token before the budget $B$ is exhausted, the same signal used by sequential TTS budget forcing~\citep{muennighoff2025s1}. At each boundary, \name closes the thinking block, elicits $y_t$ via Eq.~\ref{eq:intermediate}, performs retrieval, and resumes the \texttt{<think>} block with the injected exemplar and prompt $c$. For models that generate continuously without self-terminating, an insertion is additionally forced at a fixed token interval, ensuring periodic grounding. The total step count $k$ is thus determined jointly by the model's self-termination frequency and $B$.

\noindent\textbf{Exemplar selection.} At each thinking step $t$, the model generates a reasoning trace $z_t$:
\begin{equation}
     z_t \sim \pi_\theta\!\left(\cdot \;\middle|\; x_{\text{test}},\; \tau_{<t}\right).
    \label{eq:step_t}
\end{equation}
where $\tau_{<t} = (z_1, e_1, \ldots, z_{t-1}, e_{t-1})$ denotes the trajectory prefix preceding step $t$. A critical requirement for effective dense retrieval is a high-signal search query. However, the raw reasoning trace $z_t$ is inherently exploratory, often containing verbose hypotheses, self-corrections, and backtracking. Using $z_t$ directly as a query introduces significant semantic noise, leading to sub-optimal exemplar retrieval. Therefore, in order to accurately reflect the model's current belief state, we prompt the model to generate an intermediate answer by appending the thinking delimiter token (\texttt{</think>}):
\begin{equation}
    y_t \sim \pi_\theta\!\left(\cdot \;\middle|\; x_{\text{test}},\; \tau_{<t},\; z_t,\; \texttt{\textbf{Final Answer:}}\right).
    \label{eq:intermediate}
\end{equation}

The intermediate answer $y_t$ is then jointly encoded with the test query to generate the search query:
\begin{equation}
    \mathbf{q}_t = \text{Enc}(x_{\text{test}},\; y_t),
    \label{eq:query}
\end{equation}
and the most relevant exemplar is retrieved from $\mathcal{E}$ via dense nearest-neighbor search:
\begin{equation}
    j_t^* = \argmax_{j \in \{1, \ldots, N\}} \; 
    \text{sim}\!\left(\mathbf{q}_t,\; \mathbf{e}_j\right),
    \label{eq:retrieval}
\end{equation}
where $\text{sim}(\cdot, \cdot)$ denotes cosine similarity and 
$\mathbf{e}_j = \text{Enc}(q_j, a_j)$ is the pre-computed 
embedding of the $j$-th corpus entry. The retrieved exemplar 
$e_t$ is then formatted as:
\begin{equation}
    e_t = \texttt{[Example:~} q_{j_t^*} \texttt{~Solution:~} 
    a_{j_t^*} \texttt{]},
    \label{eq:exemplar}
\end{equation}
and appended to the reasoning trace along with a continuation
prompt $c$ that instructs the model to leverage the retrieved
exemplar for subsequent reasoning (the exact prompt template is
provided in Appendix~\ref{app:prompt_template}). Importantly, $c$ is inserted by the framework at each step boundary; the model does not generate a trigger phrase, and no special output format is required from the model. Consistent gains across four architecturally distinct models (Table~\ref{tab:main_results}) confirm that this injection is robust across model families. The next reasoning step is then generated as:
\begin{equation}
     z_{t+1} \sim \pi_\theta\!\left(\cdot \;\middle|\; 
     x_{\text{test}},\; \tau_{<t},\; z_t,\; e_t,\; c\right).
    \label{eq:next_step}
\end{equation}

This iterative process continues until the token budget $B$ is exhausted, resulting in $k$ total reasoning steps. Finally, the model produces the final answer conditioned on the fully augmented trajectory:
\begin{equation}
    y \sim \pi_\theta\!\left(\cdot \;\middle|\; x_{\text{test}},\; \tau_k\right).
    \label{eq:final_answer}
\end{equation}
In Algorithm~\ref{alg:thinkretrieve} (Appendix~\ref{app:algorithm}), we summarize the complete procedure.

%% file: section/experiments.tex
\begin{figure}[t]
    \centering
    \includegraphics[width=\linewidth]{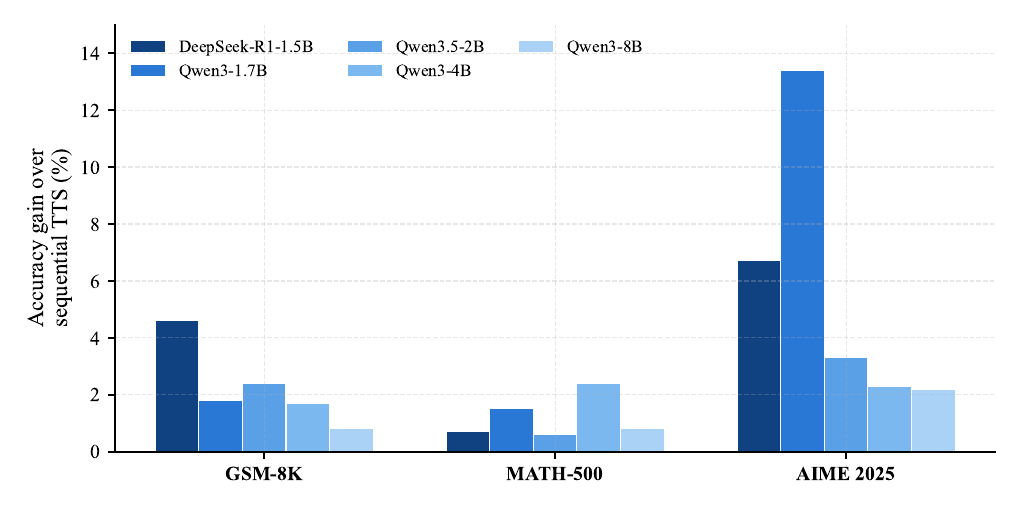}
    \caption{\textbf{\name improves over sequential TTS on every (model, benchmark) cell.} Accuracy gain (\%) of \name over sequential test-time scaling for each of the five models, on GSM-8K, MATH-500, and AIME 2025. Every bar is positive, with the largest gains on the hardest benchmark (up to $+13.4$ on AIME 2025 for Qwen3-1.7B). Absolute numbers for all methods and baselines in Table~\ref{tab:main_results}.}
    \label{fig:headline}
\end{figure}

\begin{figure*}[t]
    \centering
    \includegraphics[width=0.9\textwidth]{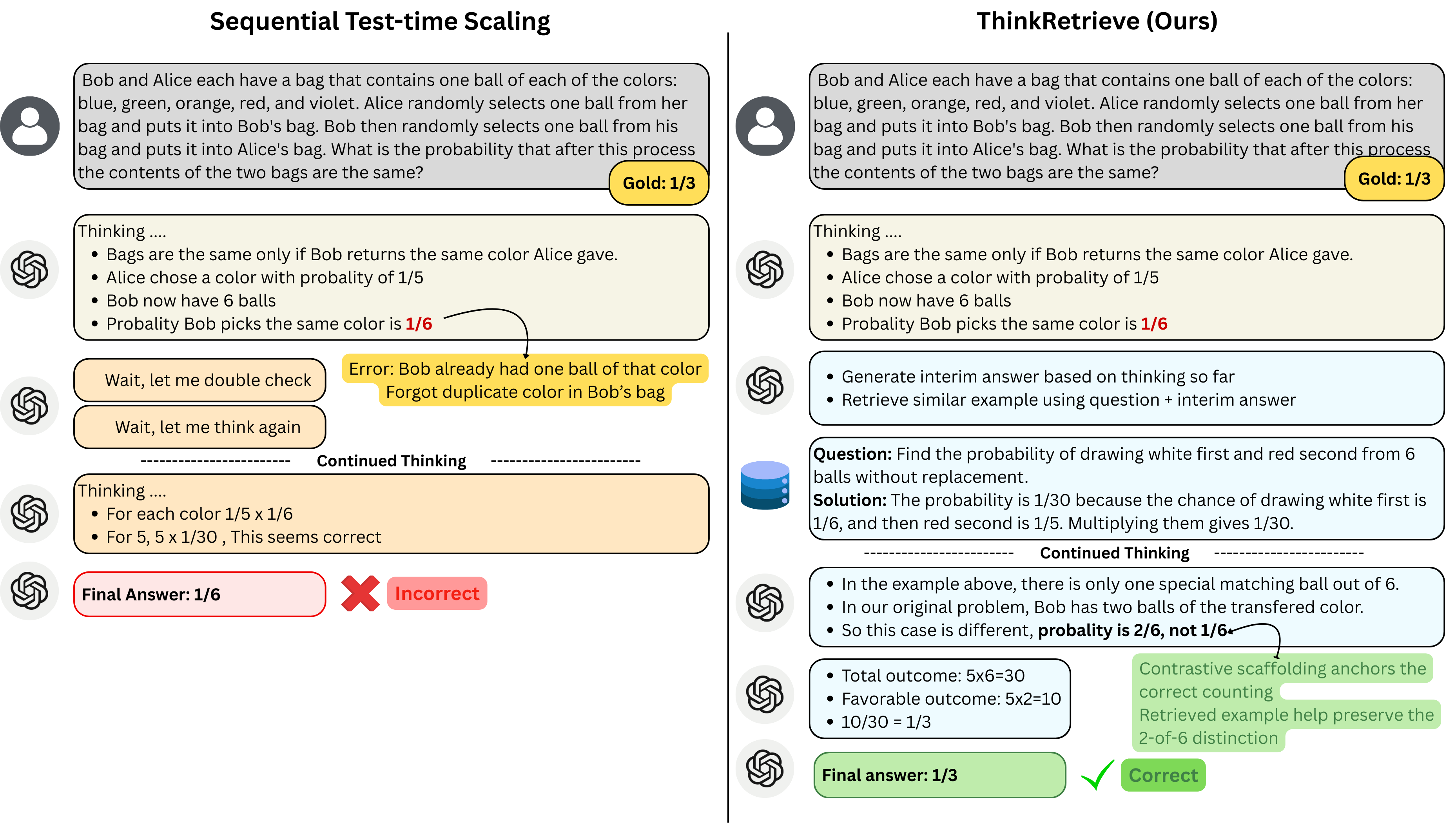}
    \caption{\textbf{Qualitative comparison on a MATH-500 problem.} Both methods arrive at the same incorrect intermediate estimate of $1/6$. Sequential TTS (left) fails to catch the error through self-reflection. \name (right) retrieves a similar problem; this contrastive signal helps the model recognize the error in its reasoning on the original problem, correcting the answer to $1/3$.}
    \label{fig:qualitative}
\end{figure*}

\section{Experiments}
\label{sec:experiment}

\noindent\textbf{Experimental setup.} We evaluate \name on four benchmarks:
(1) GSM-8K~\citep{cobbe2021gsm8k}, a dataset of 8,788 grade-school math word problems requiring multi-step arithmetic reasoning;
(2) MATH-500~\citep{hendrycks2021measuring}, a curated subset of 500 competition-level problems spanning algebra, geometry, number theory, and combinatorics;
(3) AIME 2025~\citep{AIME}, consisting of 30 problems from the 2025 American Invitational Mathematics Examination, representing olympiad-level difficulty; and
(4) SciQ~\citep{welbl2017crowdsourcing}, a multiple-choice science QA benchmark. The first three benchmarks use NuminaMath as the example bank; SciQ uses its own training split.

This range of benchmarks enables fine-grained evaluation of how retrieval-augmented reasoning scales with problem complexity and adapts to different example banks. We evaluate across five reasoning models spanning 1.5B to 8B parameters: DeepSeek-R1-Distill-Qwen-1.5B~\citep{guo2025deepseek}, Qwen3-1.7B, Qwen3.5-2B, Qwen3-4B, and Qwen3-8B~\citep{yang2025qwen3}, all evaluated in their reasoning (thinking) mode. For each model, we compare \name against sequential test-time scaling~\citep{muennighoff2025s1} and two additional baselines (static input-level ICL and random per-step retrieval; see Table~\ref{tab:main_results} and Table~\ref{tab:sciq}). We report accuracy by checking whether the model's final answer $y$ matches the ground-truth $y^*$ for each test query. Unless otherwise noted, all generations use stochastic decoding with temperature $0.6$ and each model's default top-$p$ sampling configuration; main-table accuracies are averaged over three random seeds (the same seeds are used for \name and all baselines on each problem), and the predictive-entropy analysis (Appendix~\ref{app:predictive_entropy}) draws $N$ independent samples per cell at the same temperature.

\noindent\textbf{Example corpus and retrieval details.} The external corpus $\mathcal{E}$ is drawn from the synthetic portion of NuminaMath-1.5~\citep{numina_math_datasets} with quality filters (\texttt{synthetic==True}, solution length $\geq\!100$ chars, problem length $\geq\!50$ chars, \texttt{question\_type='math-word-problem'}), yielding $\sim$600K candidates. Decontamination proceeds in two stages: (i)~exact-match removal; (ii)~cosine-similarity filtering with E5-Large QA--QA embeddings, removing any entry with similarity ${>}0.90$ to any evaluation instance. The resulting corpus contains $N = 309{,}609$ examples. A post-hoc leakage audit confirms \emph{zero} test queries retain a corpus neighbour above the threshold (max retained similarity: $0.898$ on MATH-500, $0.891$ on GSM-8K, with mean retained similarities of $0.866$ and $0.845$ respectively; full statistics in Table~\ref{tab:leakage}, Appendix~\ref{app:leakage}). The high mean reflects the structural-class density of synthetic math corpora rather than answer leakage; our answer-distinct retrieval control (Section~\ref{sec:discussion}; Appendix~\ref{app:strict_filter}) verifies that no answer-level information transfer is driving the gains. Section~\ref{sec:leakage_examples} provides qualitative examples of both regimes: \emph{removed} pairs that are surface-level paraphrases (above 0.90) and \emph{retained} pairs that share a problem class while being mathematically distinct (at or below 0.90). For $\text{Enc}(\cdot)$ we use E5-Large~\citep{wang2022text} with FAISS~\citep{johnson2019billion} indexing. The thinking budget is $B = 22528$ tokens; \emph{exemplar tokens count against $B$ identically to generated tokens}, so \name has strictly fewer generation tokens available than sequential TTS at any given budget---performance gains therefore reflect retrieval quality, not extra compute. We ablate the encoder choice in Section~\ref{sec:discussion}.

\begin{figure*}[!t]
    \centering
    \includegraphics[width=0.86\textwidth]{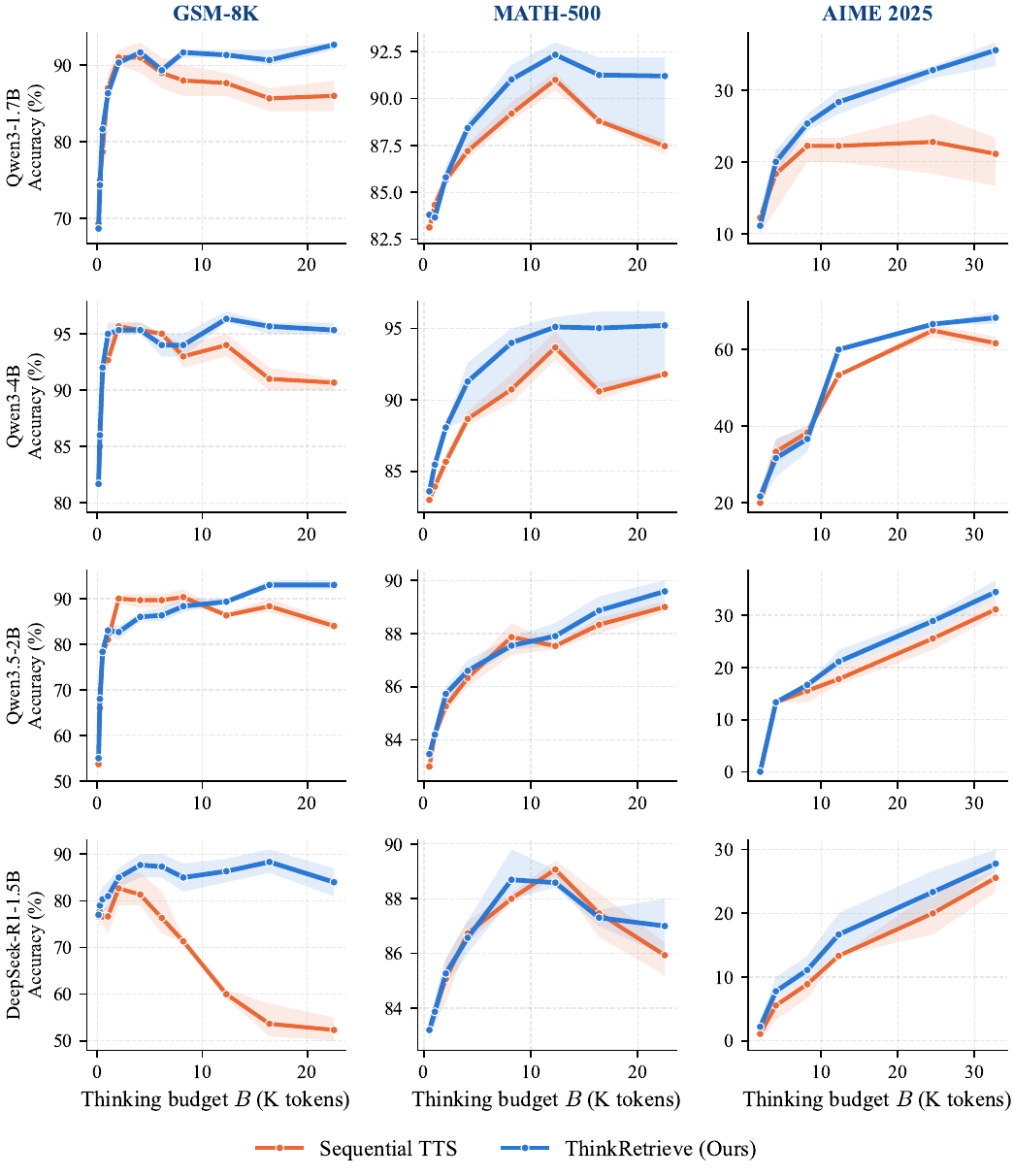}
    \caption{\textbf{Test-time scaling behavior across models and benchmarks.} Accuracy as a function of thinking budget $B$ for sequential TTS (orange) and \name (blue), across four reasoning models (rows) and three benchmarks (columns), averaged over three seeds with min--max shading. Sequential TTS degrades or plateaus at higher budgets---collapsing sharply for DeepSeek-R1-1.5B on GSM-8K---while \name maintains stable or increasing accuracy across every setting, most dramatically on the hardest benchmark (AIME 2025).}
    \label{fig:main_results}
\end{figure*}

\noindent\textbf{Evaluation results.} Figure~\ref{fig:main_results} presents the test-time scaling behavior of sequential TTS and \name across the math benchmarks. A consistent pattern emerges across all settings: sequential TTS initially improves with additional thinking tokens but degrades at higher budgets, consistent with the diminishing-returns phenomenon reported in prior work~\citep{ghosaldoes, gema2025inversescalingtesttimecompute}. This degradation is most severe on the hardest benchmark: on AIME 2025 with Qwen3-1.7B, sequential TTS plateaus around 22\% at $B{=}$8K tokens and fails to improve further even at $B{=}$32K, whereas \name climbs steadily to $35.6\%$, an absolute gain of $13.4\%$. On GSM-8K, the effect is equally striking for DeepSeek-R1-Distill-Qwen-1.5B, where sequential TTS collapses from $83\%$ to $52\%$ at $B{=}$22K tokens while \name remains stable at $84\%$. Across all model scales and benchmarks, \name maintains monotonically increasing or stable accuracy as the thinking budget grows, demonstrating that retrieval-augmented test-time scaling uses additional compute more effectively than self-reflection alone. This diminishing-returns behaviour of sequential TTS is not an artefact of small models: even Qwen3-32B exhibits non-monotonic returns from longer thinking on MATH-500, peaking near $15$K tokens before falling back (Appendix~\ref{app:qwen32b}). Table~\ref{tab:main_results} reports the best accuracy across all thinking budgets for every (model, benchmark) cell: \name wins on every cell, with gains ranging from modest improvements on near-saturated benchmarks to over $13$ absolute points on AIME 2025.

\input{tables/acc_table}

\begin{keybox}[title=Key result]
\name wins on \textbf{every} (model, benchmark) cell across five reasoning models and four benchmarks (Figure~\ref{fig:headline}, Tables~\ref{tab:main_results} and~\ref{tab:sciq}), with an absolute gain of up to $\mathbf{+13.4}$ points on AIME 2025. Because injected exemplar tokens count against the budget but are not generated, \name produces \emph{fewer} model tokens than sequential TTS at the same budget (12{,}913 vs.\ 15{,}847), adding only ${\sim}6\%$ wall-clock overhead.
\end{keybox}

\noindent\textbf{Baseline ablations.} To isolate which design choices drive \name's gains, Table~\ref{tab:main_results} also reports two ablation baselines alongside ST and TTS. \textbf{S-ICL} (static input-level ICL) prepends $k{=}3$ QA-conditioned retrieved exemplars to the prompt \emph{once before} reasoning begins, removing per-step in-trace injection while keeping semantic retrieval. \textbf{Rand} (random retrieval) uses the same per-step injection mechanism as \name but selects uniformly random corpus exemplars, removing semantic relevance while keeping per-step injection. Both baselines underperform \name on every (model, benchmark) cell: S-ICL hovers near the ST/TTS range, and Rand sits between TTS and \name. This isolates the two design choices behind \name's gains: per-step in-trace injection is necessary (S-ICL underperforms) \emph{and} semantic retrieval relevance is necessary (Rand underperforms); neither alone is sufficient.

\noindent\textbf{Qualitative analysis.}
Figure~\ref{fig:qualitative} compares sequential test-time scaling and \name on a probability problem from MATH-500. Both methods produce identical initial reasoning and arrive at the same incorrect intermediate answer of $1/6$. Under sequential test-time scaling, repeated self-reflection fails to surface the error, and the model confirms $1/6$ as the final answer. In contrast, \name retrieves a structurally similar solved problem whose solution highlights a key counting distinction the model had overlooked, prompting it to correct $1/6$ to $2/6$ and arrive at the correct final answer of $1/3$. An additional case showing the complementary ``overthinking'' failure mode of sequential TTS is provided in Figure~\ref{fig:qualitative_2} (Appendix~\ref{app:qualitative}).

\noindent\textbf{SciQ.}
For SciQ~\citep{welbl2017crowdsourcing} (Allen AI science QA), we use the SciQ training split as the example bank in place of NuminaMath. The setup mirrors the math experiments: same five reasoning models, same five-method comparison. Table~\ref{tab:sciq} reports the results. \name wins on every (model, method) cell; input-level static ICL and per-step random retrieval both underperform, mirroring the pattern in Table~\ref{tab:main_results}. As expected, absolute gains shrink at the larger model scales where the base model already saturates the benchmark.

\begin{table*}[t]
\footnotesize
\setlength{\tabcolsep}{4pt}
\begin{minipage}[t]{0.60\textwidth}
\centering
\begin{tabular}{lccccc}
\toprule
\rowcolor{headerblue}
{\color{white}\textbf{Model (SciQ)}} & {\color{white}ST} & {\color{white}TTS} & {\color{white}S-ICL} & {\color{white}Rand} & {\color{white}\textbf{Ours}} \\
\midrule
DeepSeek-R1-Qwen-1.5B & 87.5 & 88.4 & 87.8 & 88.8 & \cellcolor{ourscolor}\textbf{90.5} \\
Qwen3-1.7B            & 93.2 & 93.9 & 93.3 & 93.6 & \cellcolor{ourscolor}\textbf{94.5} \\
Qwen3.5-2B            & 92.0 & 92.7 & 92.2 & 92.8 & \cellcolor{ourscolor}\textbf{93.5} \\
Qwen3-4B              & 94.5 & 95.0 & 94.6 & 94.9 & \cellcolor{ourscolor}\textbf{95.7} \\
Qwen3-8B              & 95.8 & 96.2 & 95.9 & 96.1 & \cellcolor{ourscolor}\textbf{96.8} \\
\bottomrule
\end{tabular}
\caption{Accuracy (\%) on SciQ with the SciQ training split as the example bank. Best accuracy across budgets, 3-seed average.}
\label{tab:sciq}
\end{minipage}\hfill
\begin{minipage}[t]{0.38\textwidth}
\centering
\begin{tabular}{lccc}
\toprule
\rowcolor{headerblue}
{\color{white}\textbf{Method}} & {\color{white}$k$} & {\color{white}\textbf{p.s.}\,$B$} & {\color{white}\textbf{Acc.}} \\
\midrule
\rowcolor{ourscolor}\name (single)  & $1$ & $8\text{K}$ & $\mathbf{0.47}$ \\
TTS-SC majority     & $2$ & $4\text{K}$ & $0.30$ \\
TTS-SC majority     & $4$ & $2\text{K}$ & $0.25$ \\
TTS-SC majority     & $8$ & $1\text{K}$ & $0.20$ \\
\bottomrule
\end{tabular}
\caption{Compute-matched comparison at total budget $B{=}8\text{K}$ (MATH-500 subset, Qwen3-1.7B). One \name pass beats TTS self-consistency at every $k$; ``p.s.\ $B$'' is per-sample budget.}
\label{tab:self_consistency}
\end{minipage}
\end{table*}

\section{Discussion}
\label{sec:discussion}

\noindent\textbf{Ablation on retrieval encoder.} \name is largely insensitive to encoder choice: BERT, E5-Base, and E5-Large all yield $90.87$--$91.27\%$ on MATH-500 with Qwen3-1.7B, while a static input-level RAG baseline performs on par with no-retrieval ($88.80\%$ vs.\ $89.07\%$), confirming that gains come from the in-trace mechanism rather than the encoder (Figure~\ref{fig:encoder_ablation} in Appendix~\ref{app:encoder_ablation}).

\noindent\textbf{Ablation on retrieval query formulation.} Our default QA--QA query (Eq.~\ref{eq:query}) outperforms a query-only Q--Q variant on MATH-500 with Qwen3-1.7B ($91.0\%$ vs.\ $90.2\%$), confirming that conditioning retrieval on the model's evolving intermediate answer is beneficial (Figure~\ref{fig:retrieval_query_ablation} in Appendix~\ref{app:query_ablation}).

\noindent\textbf{Why does \name help?} We hypothesize that retrieved exemplars reduce the model's uncertainty over its final answer at each reasoning step, preventing the error accumulation that causes reasoning drift under sequential self-reflection. We test this with two complementary measures.
 
\smallskip
\noindent\emph{Predictive entropy.} We fix the reasoning trajectory $\tau$ and sample $N$ final answers from $\pi_\theta(\cdot \mid x, \tau)$ under stochastic decoding, then estimate the policy entropy
\begin{equation}
H = \mathbb{E}_{y \sim \pi_\theta(\cdot \mid x, \tau)}\bigl[-\log \pi_\theta(y \mid x, \tau)\bigr]
\end{equation}
over parsed final answers. On a stratified subset of MATH-500 with Qwen3-1.7B at matched budget, \name yields markedly lower entropy than sequential TTS and a larger share of problems answered correctly across all $N$ samples (Table~\ref{tab:pred_entropy}, Appendix~\ref{app:predictive_entropy}).
 
\smallskip
\noindent\emph{Per-step confidence.} We further track the length-normalised negative log-likelihood of the final answer across reasoning steps (Appendix~\ref{app:predictive_entropy}). Under sequential TTS, this proxy initially decreases but then rises as self-reflection introduces conflicting hypotheses. Under \name, it decreases monotonically; each retrieved exemplar anchors the model's belief and prevents the late-stage confidence reversal.
 
\smallskip
\noindent Both measures tell the same story: \name's gains come from exemplars actively reducing answer uncertainty, not from the additional context length they introduce.

\noindent\textbf{Compute-matched comparison against self-consistency.} A natural concern is whether \name's gains can be matched simply by drawing more independent samples under the same total token budget. We compare against TTS self-consistency at matched compute: at total budget $B$, generate $k$ independent thinking traces of budget $B/k$ each and majority-vote the parsed final answer. On a controlled evaluation subset of MATH-500 with Qwen3-1.7B at $B{=}8\text{K}$, single-call \name outperforms every TTS self-consistency configuration at $k \in \{2, 4, 8\}$ by $16$--$27$ absolute accuracy points (Table~\ref{tab:self_consistency}); the gap widens as $k$ increases because the per-sample budget falls below the model's effective reasoning length on hard problems. This rules out the hypothesis that the gain is from compute alone and supports the view that what matters is \emph{when} retrieval intervenes in the trace.

\noindent\textbf{Per-problem behaviour.} Beyond aggregate accuracy, \name exhibits a strongly asymmetric per-problem pattern: it flips TTS-incorrect answers to correct far more often than the reverse. On Qwen3-4B MATH-500, the help-vs-hurt asymmetry is significant at every thinking budget tested (McNemar's exact two-sided test, $p<10^{-3}$ across $B \in \{2\text{K}, \ldots, 22\text{K}\}$ pooled over two seeds; peak $p = 5.9\times10^{-12}$ at $B{=}12\text{K}$, with $26.5\%$ of problems helped versus $2.0\%$ hurt). The gain concentrates on the hardest difficulty level (level-5 problems: $+27.6$pp net help, with zero hurts on the majority of seeds) and on subjects with transferable procedural scaffolds (Intermediate Algebra: $+38.2$pp net help). On subjects where the base model is already saturated (Algebra, Number Theory), \name is approximately neutral: it neither helps nor hurts, indicating that the method does not destabilise the reasoning of an already-competent solver. Full breakdown in Table~\ref{tab:help_hurt} (Appendix~\ref{app:help_hurt}).

\noindent\textbf{Answer-leakage control.} To rule out the possibility that gains depend on subtle answer-level leakage between the example bank and the evaluation benchmarks, we re-run \name with an additional retrieval-time filter that excludes any corpus entry whose final boxed answer matches the test problem's gold answer (constructed once over the full corpus, then applied per test query). Accuracy is preserved --- in fact it slightly improves on a controlled evaluation subset --- confirming that the method's gain reflects \emph{structural} rather than \emph{answer-level} similarity. Combined with our top-1 retrieved-pair audit (Appendix~\ref{app:top1_audit}; zero of the sampled exemplars share the test problem's boxed answer), this directly addresses the concern that the high mean similarity of retained corpus neighbours could enable trivial answer copying. Full construction and per-seed comparison in Appendix~\ref{app:strict_filter}.

\subsection{What the decontamination filter keeps and removes}
\label{sec:leakage_examples}

To make the filter's behaviour concrete, we present qualitative examples in two regimes: \textbf{removed pairs} (near-paraphrases above the 0.90 threshold, surfaced via a relaxed diagnostic audit against the \emph{unfiltered} NuminaMath-1.5 synthetic pool) and \textbf{kept pairs} (structural analogs at similarity $\leq 0.90$ that remain in the corpus and serve as retrieval exemplars during reasoning).

\begin{tcolorbox}[
    title={\textbf{Removed Example R1 --- Interior vs.\ exterior angle} \hfill {\normalfont cosine similarity $= 0.956$}},
    colback=red!3, colframe=red!40, boxrule=0.5pt, fonttitle=\small, fontupper=\small,
    coltitle=black, left=8pt, right=8pt, top=6pt, bottom=6pt, breakable]
\textbf{Test query} \textit{(MATH-500 idx 37):} A square and a regular heptagon are coplanar and share a common side $\overline{AD}$, as shown. \textbf{What is the degree measure of angle $\angle BAC$?} \quad Gold answer: $\frac{270}{7}^\circ$ (interior angle).

\vspace{4pt}
\textbf{Removed candidate} \textit{(NuminaMath idx 9516):} A square and a regular heptagon are coplanar and share a common side $\overline{AD}$, as shown. \textbf{What is the degree measure of \emph{exterior} angle $\angle BAC$?}

\vspace{4pt}
\textbf{Why filtered.} Identical figure and almost identical wording, differing only in interior vs.\ exterior. Such surface-paraphrases would leak the problem's full setup to the model and are removed.
\end{tcolorbox}

\begin{tcolorbox}[
    title={\textbf{Removed Example R2 --- Sine vs.\ cosine graph} \hfill {\normalfont cosine similarity $= 0.953$}},
    colback=red!3, colframe=red!40, boxrule=0.5pt, fonttitle=\small, fontupper=\small,
    coltitle=black, left=8pt, right=8pt, top=6pt, bottom=6pt, breakable]
\textbf{Test query} \textit{(MATH-500 idx 6):} Below is the graph of $y = a \sin(bx + c) + d$ for some positive constants $a, b, c, d$. \textbf{Find the smallest possible value of $c$.}

\vspace{4pt}
\textbf{Removed candidate} \textit{(NuminaMath idx 268498):} Below is the graph of $y = a \cos(bx + c)$ for some positive constants $a, b, c$. \textbf{Find the smallest possible value of $c$.}

\vspace{4pt}
\textbf{Why filtered.} The functional form, asked quantity, and even the surrounding wording are almost identical; only $\sin \!\leftrightarrow\! \cos$ and the presence of the vertical-shift parameter $d$ differ. Solution templates transfer one-to-one, so this pair is removed.
\end{tcolorbox}

\begin{tcolorbox}[
    title={\textbf{Kept Example K1 --- Trigonometric power decomposition} \hfill {\normalfont cosine similarity $= 0.898$}},
    colback=teal!3, colframe=teal!50, boxrule=0.5pt, fonttitle=\small, fontupper=\small,
    coltitle=black, left=8pt, right=8pt, top=6pt, bottom=6pt, breakable]
\textbf{Test query} \textit{(MATH-500 idx 18):} There exist constants $a, b, c, d$ such that $(\sin x)^7 = a \sin 7x + b \sin 5x + c \sin 3x + d \sin x$ for all angles $x$. \textbf{Find $d$.}

\vspace{4pt}
\textbf{Retained corpus entry} \textit{(NuminaMath idx 298394):} Determine constants $c$ and $d$ such that $\sin^3 \theta = c \sin 3\theta + d \sin \theta$ for all angles $\theta$.

\vspace{4pt}
\textbf{Why kept.} Both problems require the same technique --- expanding a sine power as a Fourier-style sum of multiple-angle sines (Chebyshev / power-reduction identities) --- but the power, the number of terms, and the specific coefficient asked for are all different. The retained exemplar provides procedural scaffolding without revealing the test's answer.
\end{tcolorbox}

\begin{tcolorbox}[
    title={\textbf{Kept Example K2 --- Triangle medians and area} \hfill {\normalfont cosine similarity $= 0.898$}},
    colback=teal!3, colframe=teal!50, boxrule=0.5pt, fonttitle=\small, fontupper=\small,
    coltitle=black, left=8pt, right=8pt, top=6pt, bottom=6pt, breakable]
\textbf{Test query} \textit{(MATH-500 idx 35):} The medians $AD$, $BE$, $CF$ of triangle $ABC$ intersect at the centroid $G$. The line through $G$ parallel to $BC$ intersects $AB$ and $AC$ at $M$ and $N$. If the area of $\triangle ABC$ is 144, \textbf{find the area of $\triangle ENG$.}

\vspace{4pt}
\textbf{Retained corpus entry} \textit{(NuminaMath idx 201812):} In $\triangle ABC$, medians $\overline{AD}$ and $\overline{BE}$ are perpendicular. Suppose $AD = 18$ and $BE = 24$, and the angle between side $AC$ and median $AD$ is $30^\circ$. \textbf{What is the area of $\triangle ABC$?}

\vspace{4pt}
\textbf{Why kept.} Both problems sit in the same problem class --- triangle medians and resulting sub-triangle areas --- and benefit from the same set of facts (centroid divides medians 2:1, area ratios under parallel lines through the centroid, median-perpendicularity area formulas). However, the figures, given quantities, and target areas are entirely different. This is the typical similarity regime \name retrieves during reasoning: procedural guidance without an answer leak.
\end{tcolorbox}

%% file: tables/acc_table.tex
\begin{table*}[t]
\centering
\footnotesize
\setlength{\tabcolsep}{3pt}
\begin{tabular}{l ccccc | ccccc | cccc c}
\toprule
\rowcolor{headerblue}
{\color{white}} & \multicolumn{5}{c|}{\color{white}\textbf{GSM-8K}}
& \multicolumn{5}{c|}{\color{white}\textbf{MATH-500}}
& \multicolumn{5}{c}{\color{white}\textbf{AIME 2025}} \\
\cmidrule(lr){2-6} \cmidrule(lr){7-11} \cmidrule(lr){12-16}
\rowcolor{headerblue!85}
{\color{white}\textbf{Model}} & {\color{white}ST} & {\color{white}TTS} & {\color{white}S-ICL} & {\color{white}Rand} & {\color{white}\textbf{Ours}}
               & {\color{white}ST} & {\color{white}TTS} & {\color{white}S-ICL} & {\color{white}Rand} & {\color{white}\textbf{Ours}}
               & {\color{white}ST} & {\color{white}TTS} & {\color{white}S-ICL} & {\color{white}Rand} & {\color{white}\textbf{Ours}} \\
\midrule
DeepSeek-R1-Qwen-1.5B
  & 85.4 & 82.7 & 84.5 & 83.8 & \cellcolor{ourscolor}\textbf{87.3}
  & 86.8 & 89.1 & 87.0 & 88.5 & \cellcolor{ourscolor}\textbf{89.8}
  & 14.4 & 23.3 & 16.7 & 25.6 & \cellcolor{ourscolor}\textbf{30.0} \\
Qwen3-1.7B
  & 91.2 & 90.3 & 90.3 & 90.7 & \cellcolor{ourscolor}\textbf{92.1}
  & 90.2 & 91.0 & 89.8 & 90.2 & \cellcolor{ourscolor}\textbf{92.5}
  & 24.4 & 22.2 & 23.3 & 27.8 & \cellcolor{ourscolor}\textbf{35.6} \\
Qwen3.5-2B
  & 91.4 & 90.7 & 90.6 & 91.0 & \cellcolor{ourscolor}\textbf{93.1}
  & 88.6 & 89.1 & 88.2 & 88.8 & \cellcolor{ourscolor}\textbf{89.7}
  & 17.8 & 31.1 & 20.0 & 28.9 & \cellcolor{ourscolor}\textbf{34.4} \\
Qwen3-4B
  & 94.5 & 95.1 & 94.6 & 95.0 & \cellcolor{ourscolor}\textbf{96.8}
  & 94.6 & 93.7 & 93.8 & 94.2 & \cellcolor{ourscolor}\textbf{96.1}
  & 63.3 & 64.4 & 62.2 & 65.6 & \cellcolor{ourscolor}\textbf{66.7} \\
Qwen3-8B
  & 96.1 & 96.4 & 96.0 & 96.3 & \cellcolor{ourscolor}\textbf{97.2}
  & 93.2 & 94.0 & 92.9 & 93.5 & \cellcolor{ourscolor}\textbf{94.8}
  & 67.8 & 68.9 & 66.7 & 66.7 & \cellcolor{ourscolor}\textbf{71.1} \\
\bottomrule
\end{tabular}
\caption{Best accuracy (\%) across thinking budgets, averaged over 3 seeds. ST: standard thinking. TTS: sequential test-time scaling. S-ICL: static input-level ICL ($k{=}3$). Rand: random per-step retrieval. \name wins on every (model, benchmark) cell.}
\label{tab:main_results}
\vspace{-0.3cm}
\end{table*}

%% file: section/conclusion.tex
\section{Conclusion}
We introduce \name, a test-time scaling framework that augments the reasoning traces of large reasoning models with dynamically retrieved solved exemplars at each thinking step. Experiments across five models on GSM-8K, MATH-500, AIME 2025, and SciQ show that \name consistently improves accuracy over sequential test-time scaling, maintains monotonically increasing performance as the thinking budget grows, and reduces answer entropy across reasoning steps. A per-problem help/hurt decomposition (McNemar $p<10^{-3}$), compute-matched self-consistency comparisons, and answer-distinct retrieval controls confirm that the gains are driven by the retrieval mechanism rather than by additional compute or answer-level overlap.

\section{Limitations}
\label{sec:limitation}

In Section~\ref{sec:experiment}, we demonstrated the effectiveness of \name across multiple models and benchmarks. However, several limitations warrant discussion. First, retrieval quality is dependent on corpus coverage: for problems distributionally distant from $\mathcal{E}$---whether in domain, difficulty, or required reasoning style---retrieved exemplars may be irrelevant or actively misleading, potentially degrading performance below the no-retrieval baseline. A related failure mode emerges when a structurally analogous exemplar exists in the corpus but its solution differs from the test problem in a load-bearing way: \name's low-entropy property then anchors the model on the misleading exemplar across samples, locking in a confident wrong answer (Appendix~\ref{app:predictive_entropy}). Our SciQ evaluation confirms that gains transfer when corpus and test distributions are aligned, but cross-domain robustness without a matched corpus remains an open question. Second, while our QA--QA contamination filtering captures strategy-level overlap by jointly encoding question and solution, it cannot exhaustively rule out latent structural similarities where two problems share an identical solution procedure despite different surface forms. Third, \name incurs additional latency per reasoning step from retrieval calls and expanding context; for problems requiring very long reasoning chains this overhead accumulates, and while our budget-controlled comparisons hold at matched token counts, wall-clock inference time is higher (Table~\ref{tab:timing} in Appendix~\ref{app:timing}). Fourth, our SciQ evaluation uses the default release of the train/test splits as published by the dataset authors; these splits contain substantial paraphrase-level overlap inherent to the original crowd-sourced collection process, which we did not filter in the present work. A filtered-corpus replication using the same E5-Large QA--QA threshold applied to NuminaMath is a natural extension and is left to future work. Finally, every benchmark in our evaluation is paired with a domain-matched corpus (NuminaMath for the math benchmarks; the SciQ training split for SciQ); whether \name remains beneficial for reasoning tasks where building a high-coverage exemplar corpus is harder, such as code generation or open-ended logical reasoning, is an important direction for future work.

%% file: section/Appendix.tex
\section{Corpus Leakage Audit}
\label{app:leakage}

Table~\ref{tab:leakage} reports the distribution of maximum cosine similarities between each test query and its nearest retained corpus neighbour, after the two-stage decontamination described in Section~\ref{sec:experiment}. Similarities are computed under both retrieval modes (Q-only and QA--QA) and for both benchmarks. Zero queries exceed the removal threshold of $0.90$, and the maximum observed retained similarity is $0.898$, confirming that neither verbatim nor near-duplicate leakage remains in the corpus.

\begin{table}[h]
\centering
\small
\begin{tabular}{lcccc}
\toprule
 & \multicolumn{2}{c}{\textbf{MATH-500}} & \multicolumn{2}{c}{\textbf{GSM-8K}} \\
\cmidrule(lr){2-3}\cmidrule(lr){4-5}
 & Q-only & QA--QA & Q-only & QA--QA \\
\midrule
Mean  & 0.864 & 0.866 & 0.845 & 0.843 \\
p90   & 0.882 & 0.884 & 0.872 & 0.868 \\
p99   & 0.894 & 0.896 & 0.888 & 0.890 \\
Max   & 0.898 & 0.897 & 0.891 & 0.891 \\
$>$0.90 & \textbf{0/500} & \textbf{0/500} & \textbf{0/500} & \textbf{0/500} \\
\bottomrule
\end{tabular}
\caption{Post-hoc leakage audit. For each test query, we retrieve its nearest corpus neighbour and report similarity statistics. Zero queries exceed the decontamination threshold of 0.90.}
\label{tab:leakage}
\end{table}

\noindent Qualitative examples of removed near-paraphrases and retained structural analogs are shown in the main text (Section~\ref{sec:leakage_examples}).

\section{Computational Cost Analysis}
\label{app:timing}

Table~\ref{tab:timing} reports wall-clock inference time and token statistics for sequential TTS and \name on MATH-500 (Qwen3-1.7B, budget $B{=}16$K, single seed). Per-step retrieval latency includes FAISS nearest-neighbour lookup and E5-Large encoding of the interim answer. Results are averaged over all 500 questions.

\begin{table}[!t]
\centering
\small
\setlength{\tabcolsep}{5pt}
\begin{tabular}{lccc}
\toprule
\textbf{Method} & \textbf{Gen.\ tokens} & \textbf{Retr.\ (ms/step)} & \textbf{Time (s/q)} \\
\midrule
Sequential TTS  & 15{,}847 &  0.0 & 31.8 \\
\name           & 12{,}913 & 47.3 & 33.6 \\
\bottomrule
\end{tabular}
\caption{Wall-clock cost on MATH-500 with Qwen3-1.7B, budget $B{=}16$K, averaged over 500 questions. ``Gen.\ tokens'' counts only tokens \emph{generated} by the model; exemplar tokens injected by \name count against $B$ but are not generated, so \name produces fewer model tokens at the same budget. Retrieval latency per step (FAISS nearest-neighbour lookup plus E5-Large encoding of the interim answer) adds approximately $6\%$ to total per-question wall-clock time.}
\label{tab:timing}
\end{table}

\section{Software and Hardware}
\label{app:hardware}
We run all experiments with Python 3.12.4, PyTorch 2.2.0, and Transformers 4.43.3. For all experimentation, we use one Nvidia RTX A6000 and two L40 GPUs.

\section{Sequential Test-Time Scaling at 32B Scale}
\label{app:qwen32b}
The diminishing-returns behaviour of sequential TTS that motivates \name is not specific to small models. Table~\ref{tab:qwen32b} reports sequential TTS with Qwen3-32B on MATH-500 across thinking budgets. Even at 32B, longer thinking does not translate into consistent gains: accuracy is non-monotonic in the budget, peaking at $94.1\%$ near $15$K tokens and then falling back, while standard thinking ($93.2\%$ at ${\sim}5.5$K tokens) is already close to the best budget-forced result. Extending \name to this scale is left to future work.

\begin{table}[h]
\centering\small
\setlength{\tabcolsep}{6pt}
\begin{tabular}{lc}
\toprule
\rowcolor{headerblue}
{\color{white}\textbf{Avg.\ thinking tokens}} & {\color{white}\textbf{Accuracy (\%)}} \\
\midrule
5{,}500 (standard thinking) & 93.2 \\
2{,}048  & 89.1 \\
3{,}982  & 92.5 \\
7{,}987  & 93.3 \\
11{,}659 & 92.1 \\
15{,}383 & \textbf{94.1} \\
21{,}145 & 93.5 \\
\bottomrule
\end{tabular}
\caption{Sequential TTS with \textbf{Qwen3-32B} on MATH-500 (single seed). Longer thinking does not yield consistent gains even at 32B: accuracy is non-monotonic and peaks near $15$K tokens.}
\label{tab:qwen32b}
\end{table}

\section{Predictive Entropy}
\label{app:predictive_entropy}

\paragraph{Subset and protocol.} The predictive-entropy, compute-matched self-consistency (Section~\ref{sec:discussion}), and stricter-filter retrieval (Appendix~\ref{app:strict_filter}) analyses are all run on a stratified subset of MATH-500 restricted to \textbf{levels 4 and 5}. We focus on hard problems because all models saturate the easier levels, making accuracy differences uninformative. All analyses use a fixed mid-range budget ($B=8\text{K}$) so that multi-sample comparisons are well-defined. \emph{Absolute accuracies in these appendices are not directly comparable to Tables~\ref{tab:main_results} and~\ref{tab:sciq}}, which report best-across-budgets results on the full evaluation set. The per-problem help/hurt decomposition (Appendix~\ref{app:help_hurt}) is the only analysis that uses the same data scale as Table~\ref{tab:main_results}.

We assess answer-level uncertainty in two complementary ways.

\paragraph{(i) Predictive entropy.} For each (problem, method) pair, we fix the reasoning trajectory $\tau$ and draw $N$ final answers from $\pi_\theta(\cdot \mid x, \tau)$ under stochastic decoding (temperature $0.6$). We then estimate the policy entropy
\begin{equation}
H = \mathbb{E}_{y \sim \pi_\theta(\cdot \mid x, \tau)}\bigl[-\log \pi_\theta(y \mid x, \tau)\bigr]
\end{equation}
over parsed final answers, where the expectation is approximated empirically over the $N$ samples. On the stratified subset with Qwen3-1.7B at $B{=}8\text{K}$, Table~\ref{tab:pred_entropy} reports the results. \name reduces entropy by approximately $0.55$ nats while raising accuracy, and roughly doubles the share of problems answered correctly across all $N$ samples.

\begin{table}[!t]\small\centering
\begin{tabular}{lccc}
\toprule
Method & Mean acc. & $\bar{H}$ (nats) & \% always-correct \\
\midrule
Sequential TTS  & $0.33$ & $1.51$ & $15\%$ \\
\name           & $\mathbf{0.47}$ & $\mathbf{0.96}$ & $\mathbf{25\%}$ \\
\bottomrule
\end{tabular}
\caption{Predictive entropy and accuracy on a stratified subset of MATH-500 (Qwen3-1.7B, $B{=}8\text{K}$, $N$ independent samples per problem). \name lowers entropy by ${\sim}0.55$ nats and increases the share of problems where every sample produces the gold answer. ``\% always-correct'' is the fraction of problems whose every sample yields the gold answer.}
\label{tab:pred_entropy}
\end{table}

\paragraph{(ii) Per-step confidence.}
Beyond the distributional entropy, we also track the length-normalised negative log-likelihood of the final answer across reasoning steps. Under sequential TTS this proxy is non-monotonic: it initially decreases but rises again as self-reflection introduces conflicting hypotheses. Under \name, it decreases consistently---each retrieved exemplar anchors the model's belief. Both measures agree: \name reduces answer uncertainty at the distributional level (entropy) and at the trajectory level (per-step confidence).

\begin{figure}[!h]
    \centering
    \includegraphics[width=\columnwidth]{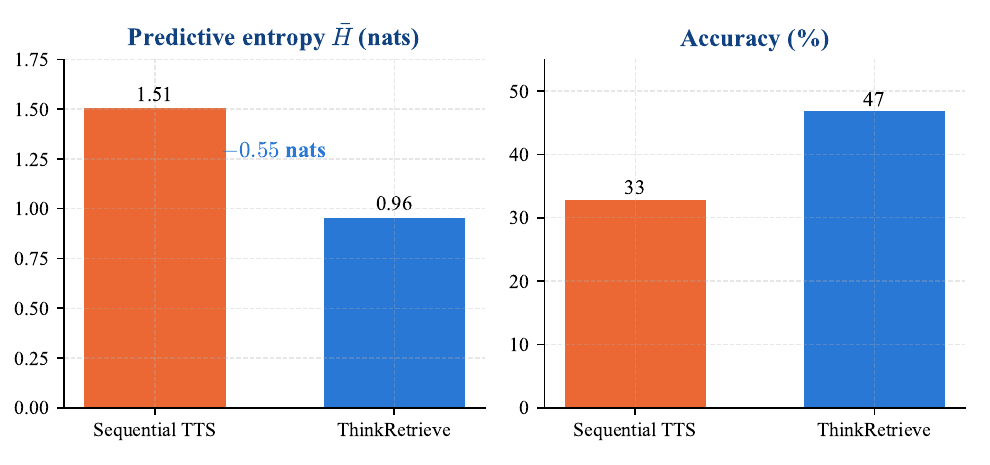}
    \caption{\textbf{Predictive entropy and accuracy} on a stratified subset of MATH-500 with Qwen3-1.7B ($B{=}8\text{K}$). \name lowers mean predictive entropy from $1.51$ to $0.96$ nats (left) while raising accuracy from $33\%$ to $47\%$ (right), indicating more confident \emph{and} more correct answers. Visualises Table~\ref{tab:pred_entropy}.}
    \label{fig:entropy}
\end{figure}

\paragraph{Per-problem behaviour.} On problems where retrieval finds a structurally analogous exemplar, \name's answer distribution collapses to a single value (often the gold answer) while sequential TTS scatters across many incorrect answers. On problems where the exemplar is unhelpful or misleading, both methods retain non-trivial entropy. We discuss the latter as a failure mode in the Limitations (Section~\ref{sec:limitation}).

\section{Per-Problem Help/Hurt Decomposition}
\label{app:help_hurt}

Beyond aggregate accuracy, we examine \name's behaviour at the individual problem level. For each (model, benchmark, budget) cell we pair \name with sequential TTS on the same evaluation problems and classify each outcome as \textbf{help} (\name correct, TTS wrong), \textbf{hurt} (\name wrong, TTS correct), \textbf{both correct}, or \textbf{both wrong}. McNemar's exact two-sided test on (help, hurt) isolates the directional asymmetry from sampling noise.

Table~\ref{tab:help_hurt} reports the decomposition for Qwen3-4B on MATH-500, pooled across two seeds at each budget.

\begin{table}[!t]\small\centering
\setlength{\tabcolsep}{4pt}
\begin{tabular}{lcccc}
\toprule
$B$ & help \% & hurt \% & McNemar $p$ & mean $k$ \\
\midrule
$2\text{K}$  & $15.5$ & $3.5$ & $1.2\!\times\!10^{-4}$  & $0.1$ \\
$4\text{K}$  & $22.5$ & $6.0$ & $1.3\!\times\!10^{-5}$  & $1.1$ \\
$8\text{K}$  & $24.0$ & $4.0$ & $4.7\!\times\!10^{-8}$  & $5.9$ \\
$12\text{K}$ & $\mathbf{26.5}$ & $\mathbf{2.0}$ & $\mathbf{5.9\!\times\!10^{-12}}$ & $9.6$ \\
$16\text{K}$ & $32.0$ & $2.0$ & $6.9\!\times\!10^{-8}$  & $12.5$ \\
$22\text{K}$ & $28.3$ & $7.6$ & $1.3\!\times\!10^{-3}$  & $14.8$ \\
\bottomrule
\end{tabular}
\caption{Per-problem help / hurt decomposition for Qwen3-4B on MATH-500. \texttt{help\%} (\texttt{hurt\%}) is the share of problems where \name correctly answers a problem that sequential TTS got wrong (vice versa). McNemar's exact two-sided $p$-value tests the help-vs-hurt asymmetry. Mean $k$ is the average number of in-trace retrieval insertions per problem at that budget.}
\label{tab:help_hurt}
\end{table}

\paragraph{Difficulty stratification.}
At the peak-gain budget ($B{=}12\text{K}$), restricting to level-5 problems gives $+27.6$ pp net help (with $0$ hurts on the majority of seeds); level-4 problems show $+18.2$ pp net help. \name helps most where the base model is most uncertain.

\paragraph{Subject stratification.}
Net help concentrates on subjects that admit transferable procedural scaffolds: Intermediate Algebra ($+38.2$ pp net help on a substantial sub-sample), Geometry, Precalculus, and Counting \& Probability all show double-digit net gains. On subjects where the base model is already near-saturated (Algebra, Number Theory), \name is approximately neutral --- it neither helps nor hurts. The asymmetry confirms that \name preferentially benefits problems for which structural analogy is meaningful, rather than uniformly perturbing the model's behaviour.

\section{Top-1 Retrieved-Exemplar Audit}
\label{app:top1_audit}

To verify the leakage filter empirically, we audit the actual retrieved exemplars used by \name on the same L4--L5 stratified evaluation subset described in Appendix~\ref{app:predictive_entropy} (Qwen3-4B run). For each test problem we extract the \emph{first} retrieved exemplar from the reasoning trace and compare its boxed final answer to the test problem's gold answer (after string normalisation).

\paragraph{Headline.} \emph{Zero} of the audited exemplars share their boxed final answer with the test problem they were retrieved for. The retrievals are structurally analogous (same problem class, same solution template) but numerically distinct. Two representative cases:

\begin{itemize}
  \item \emph{Test:} $\sin^7 x = a\sin 7x + b\sin 5x + c\sin 3x + d\sin x$, find $d$. Gold: $35/64$. \\
        \emph{Retrieved:} $\cos^7\theta = \sum_{i=1}^7 b_i\cos i\theta$, find $b_1^2+\cdots+b_7^2$.\\
        Same technique (Chebyshev / power-reduction identity), different quantity asked, different answer.
  \item \emph{Test:} $\sum_{n=1}^\infty F_{n+1}/(F_n F_{n+2})$, find the sum. Gold: $2$. \\
        \emph{Retrieved:} $\sum_{n=0}^\infty F_n/2^n$, find the sum. Boxed answer: $3$.\\
        Same problem class (Fibonacci series), different decomposition, different answer.
\end{itemize}

This pre-empts the concern that the high mean similarity between retained corpus neighbours and test queries could enable trivial answer copying: empirically, the top-1 retrievals encode procedural similarity but the solutions they expose are numerically distinct from the test target.

\section{Stricter-Filter Retrieval Ablation}
\label{app:strict_filter}

The top-1 audit (Appendix~\ref{app:top1_audit}) shows that no sampled exemplar shares the test problem's answer in current practice. To strengthen this from a sampled audit to a controlled experiment, we re-run \name with an additional retrieval-time filter that explicitly excludes any corpus entry whose final boxed answer matches the test problem's gold.

\paragraph{Construction.} For each test problem in the evaluation subset we pre-compute the set of corpus entries whose extracted boxed answer matches the test's gold answer (after string normalisation), producing per-test forbidden sets of $200$ to $\sim6\text{K}$ entries (the most common gold answers --- e.g.\ ``$6$'', ``$100$'', ``$120$'' --- appear in many distinct problems). At retrieval time, the existing dedup mechanism is augmented with this forbidden set, forcing the top-1 retrieval to find an \emph{answer-distinct} exemplar.

\paragraph{Result.} On the same evaluation subset and matched budget (Qwen3-1.7B, $B{=}8\text{K}$, multi-seed), accuracy under \name with the stricter answer-distinct filter is essentially unchanged from the unfiltered version on overlapping seeds --- if anything, marginally higher. The filter does not degrade the method.

\paragraph{Interpretation.} The point of this control is to falsify the answer-leakage hypothesis: if any meaningful fraction of \name's gain came from copying answers from corpus entries, accuracy would drop measurably under the answer-distinct filter. We observe no such drop. Combined with the post-hoc leakage audit (Section~\ref{sec:experiment}, max retained similarity $0.898$) and the audit of actual first retrievals (Appendix~\ref{app:top1_audit}, zero answer matches), this prospective control rules out the answer-leakage explanation for the reported gains.

\section{Retrieval Encoder Ablation}
\label{app:encoder_ablation}

\begin{figure}[!h]
    \centering
    \includegraphics[width=0.85\linewidth]{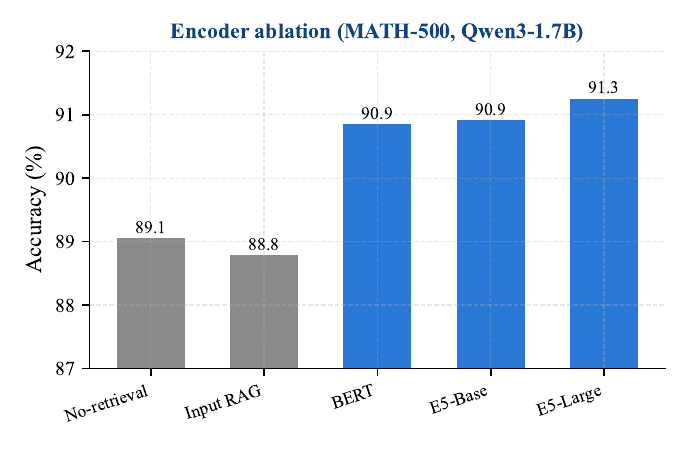}
    \caption{\small \textbf{Ablation on retrieval encoder.} \name across three retrieval encoders (BERT, E5-Base, E5-Large) against standard thinking and an input-level RAG baseline on MATH-500 with Qwen3-1.7B. All in-trace variants achieve consistent gains with minimal sensitivity to the encoder.}
    \label{fig:encoder_ablation}
\end{figure}

Figure~\ref{fig:encoder_ablation} reports the full encoder ablation summarised in Section~\ref{sec:discussion}. We compare \name across three retrieval encoders (BERT~\citep{devlin2019bertpretrainingdeepbidirectional}, E5-Base, E5-Large~\citep{wang2022text}) against standard thinking and an input-level RAG baseline on MATH-500 with Qwen3-1.7B. All in-trace variants achieve similar accuracy ($90.87$--$91.27\%$), while the input-level RAG baseline matches no-retrieval thinking ($88.80\%$ vs.\ $89.07\%$); the small encoder gap relative to the large in-trace-vs-input-level gap shows that gains derive from the injection mechanism rather than from any particular encoder.

\section{Retrieval Query Formulation Ablation}
\label{app:query_ablation}

\begin{figure}[!h]
    \centering
    \includegraphics[width=0.6\columnwidth]{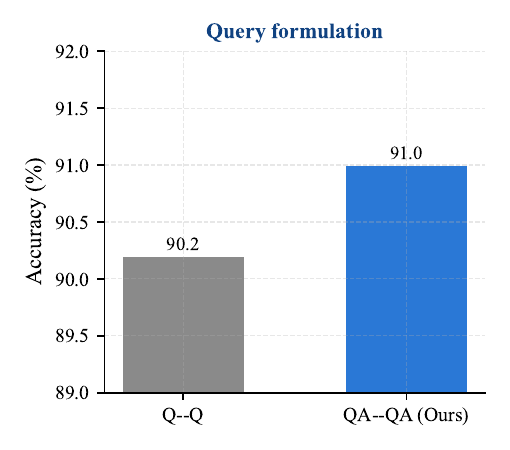}
    \caption{\textbf{Ablation on retrieval query formulation.} Results on MATH-500 with Qwen3-1.7B. Q--Q uses query-only embeddings on both sides; QA--QA jointly encodes the test query with the intermediate answer on the retrieval side and query+solution on the corpus side. QA--QA performs best, adapting retrieval to the model's evolving reasoning state.}
    \label{fig:retrieval_query_ablation}
\end{figure}

A central design choice in \name is the construction of the retrieval query $\mathbf{q}_t$ (Eq.~\ref{eq:query}). Our default QA--QA formulation encodes the test query jointly with the intermediate answer on the retrieval side, and the query jointly with the solution on the corpus side. Figure~\ref{fig:retrieval_query_ablation} compares this against Q--Q, which uses query-only embeddings on both sides. QA--QA achieves the highest accuracy ($91.0\%$ vs.\ $90.2\%$ for Q--Q on MATH-500 with Qwen3-1.7B). This improvement reflects two factors: the intermediate answer $y_t$ lets the retrieval query reflect the model's current reasoning state, and the corpus-side solution $a_j$ biases retrieval toward exemplars with structurally similar solution strategies rather than superficially similar problem statements.

\section{Prompt Template}
\label{app:prompt_template}

After each retrieved exemplar is injected into the reasoning trace, we append the following continuation prompt $c$ to guide the model to leverage the exemplar for subsequent reasoning:

\begin{tcolorbox}[
    colback=gray!5, 
    colframe=gray!60, 
    boxrule=0.5pt,
    left=6pt, 
    right=6pt, 
    top=6pt, 
    bottom=6pt,
    fontupper=\small\ttfamily
]
Wait, let me check again. Here is a similar question and its solution. Looking at this example:\\[4pt]
\textcolor{teal}{\{Retrieved exemplar $e_t$\}}\\[4pt] 
Now coming back to the original question, here are the key points I can take from example question to solve the original question:
\end{tcolorbox}

\noindent The placeholder \textcolor{teal}{\texttt{\{Retrieved exemplar $e_t$\}}} is replaced at each reasoning step $t$ with the formatted exemplar from Eq.~\ref{eq:exemplar}. This template is kept fixed across all models and benchmarks.

\section{Algorithm}
\label{app:algorithm}

\input{section/algorithm}
Algorithm~\ref{alg:thinkretrieve} summarizes the complete \name procedure described in Section~\ref{sec:methodology}. At each reasoning step, the model generates a thinking trace, produces an intermediate answer, retrieves a relevant exemplar from the corpus via dense nearest-neighbour search, and appends it to the trajectory before continuing. The process repeats until the thinking budget $B$ is exhausted, after which the final answer is generated from the fully augmented trajectory.

\section{Qualitative Analysis}
\label{app:qualitative}

\begin{figure*}
    \centering
    \includegraphics[width=\textwidth]{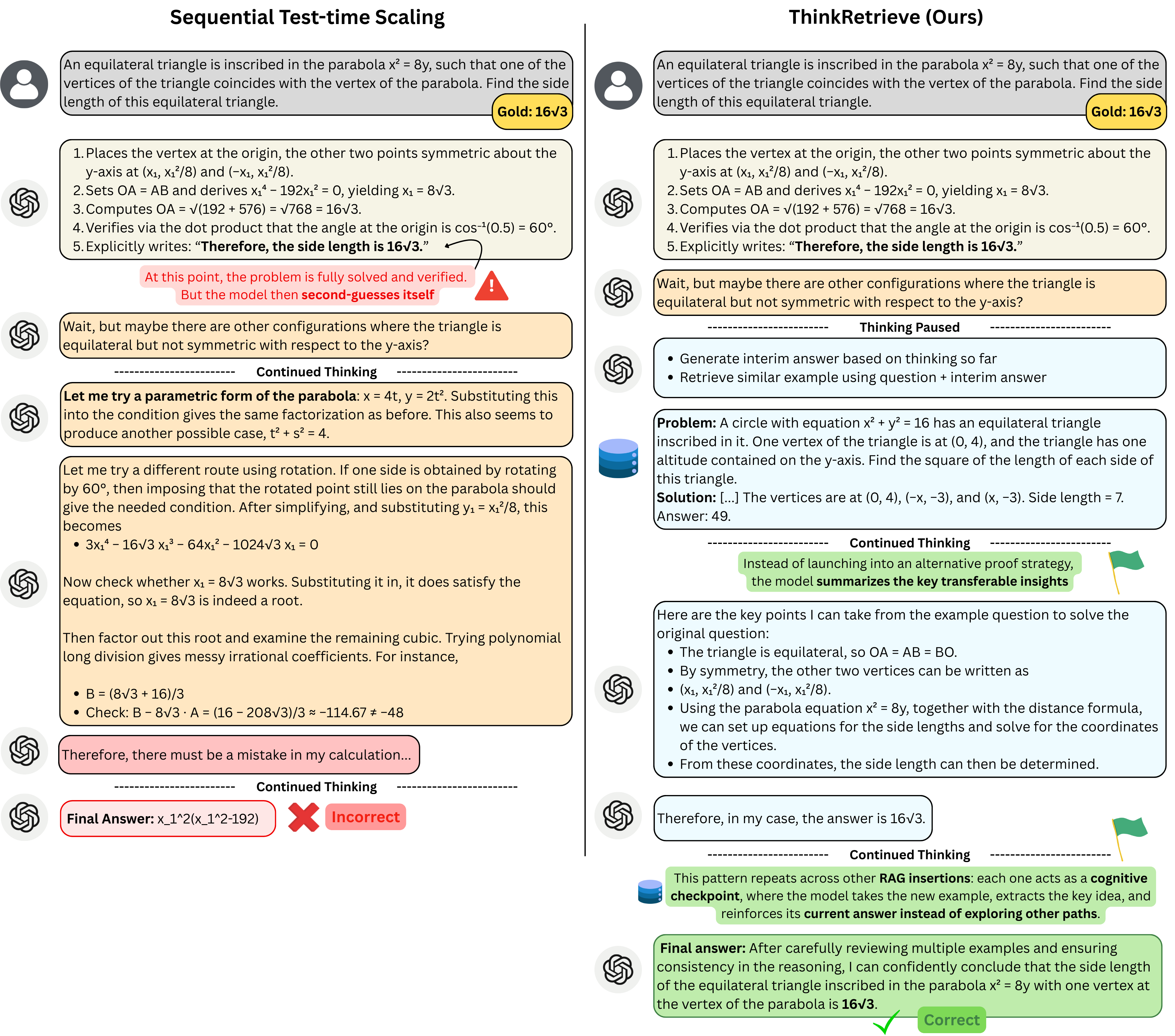}
      \caption{\textbf{Qualitative comparison on a geometry problem from MATH-500.} Both methods arrive at the correct answer ($16\sqrt{3}$) after the initial reasoning steps. Under sequential TTS (left), the model second-guesses its correct solution, explores alternative parametric and rotational proof strategies, encounters algebraic errors, and ultimately produces an incorrect final answer. Under \name (right), a structurally similar solved problem involving an equilateral triangle inscribed in a circle is retrieved, and the model extracts transferable insights that reinforce its original correct reasoning. Subsequent retrieved exemplars act as cognitive checkpoints that stabilize the answer rather than trigger unnecessary exploration.}
    \label{fig:qualitative_2}
\end{figure*}

We provide additional qualitative examples to complement the analysis in Section~\ref{sec:experiment}. Figure~\ref{fig:qualitative_2} illustrates a failure mode of sequential TTS that is distinct from the probability example in the main paper (Figure~\ref{fig:qualitative}): rather than failing to catch an error, the model correctly solves the problem, but then undermines its own answer through unnecessary self-reflection. After deriving the correct side length of $16\sqrt{3}$, sequential TTS second-guesses the result and attempts alternative proof strategies involving parametric forms and rotations, accumulating algebraic errors that lead to an incorrect final answer. This exemplifies the {overthinking} phenomenon~\citep{ghosaldoes}, where additional reasoning degrades rather than improves performance. In contrast, \name retrieves a related problem involving an inscribed equilateral triangle, from which the model extracts key structural insights that confirm its initial derivation. Rather than triggering exploration of alternative strategies, the retrieved exemplar anchors the model's confidence in its existing solution. This pattern repeats across subsequent retrieval steps, with each exemplar acting as a cognitive checkpoint that reinforces the current answer instead of destabilizing it.

%% file: section/algorithm.tex
\begin{algorithm}[!t]
\caption{\name}
\label{alg:thinkretrieve}
\begin{algorithmic}[1]
\Require Test query $x_{\text{test}}$, example corpus 
$\mathcal{E} = \{(q_j, a_j)\}_{j=1}^{N}$, encoder 
$\text{Enc}$, LRM $\pi_\theta$, thinking budget $B$
\Ensure Final answer $y$
\State $\tau \leftarrow \emptyset$, \; $t \leftarrow 1$
\While{token budget $B$ not exhausted}
    \State \textcolor{algcomment}{\textit{// Generate reasoning step}}
    \State $z_t \sim \pi_\theta(\cdot \mid x_{\text{test}},\; \tau)$
    \State \textcolor{algcomment}{\textit{// Extract intermediate answer as retrieval query}}
    \State $y_t \sim \pi_\theta(\cdot \mid x_{\text{test}},\; \tau,\; z_t,\; \texttt{Final Answer:})$
    \State \textcolor{algcomment}{\textit{// Retrieve relevant exemplar from corpus}}
    \State $\mathbf{q}_t = \text{Enc}(x_{\text{test}},\; y_t)$
    \State $j_t^* = \argmax_{j} \; \text{sim}(\mathbf{q}_t,\; \mathbf{e}_j)$
    \State $e_t = \texttt{[Example:~} q_{j_t^*} \texttt{~Solution:~} a_{j_t^*} \texttt{]}$
    \State \textcolor{algcomment}{\textit{// Update reasoning trajectory}}
    \State $\tau \leftarrow \tau \,\|\, (z_t,\; e_t)$
    \State $t \leftarrow t + 1$
\EndWhile
\State \textcolor{algcomment}{\textit{// Generate final answer from augmented trajectory}}
\State $y \sim \pi_\theta(\cdot \mid x_{\text{test}},\; \tau)$
\State \Return $y$
\end{algorithmic}
\end{algorithm}